\documentclass[conference]{IEEEtran}
\IEEEoverridecommandlockouts
\usepackage{cite}
\usepackage{amsmath,amssymb,amsfonts}
\usepackage{algorithmic}
\usepackage{graphicx}
\usepackage{textcomp}
\usepackage{multirow}
\usepackage{xcolor}
\usepackage{float}
\def\BibTeX{{\rm B\kern-.05em{\sc i\kern-.025em b}\kern-.08em
    T\kern-.1667em\lower.7ex\hbox{E}\kern-.125emX}}

\usepackage[numbers]{natbib}
\usepackage{booktabs}

\newcommand\blfootnote[1]{%
  \begingroup
  \renewcommand\thefootnote{}\footnote{#1}%
  \addtocounter{footnote}{-1}%
  \endgroup
}

\begin{document}

\title{Legal Element-oriented Modeling with Multi-view Contrastive Learning for Legal Case Retrieval}

\author{
 \IEEEauthorblockN{Wang, Zhaowei$^{1, 2}$*}
 \IEEEauthorblockA{$^{1}$\textit{Department of Computer Science and Engineering} \\
\textit{The Hong Kong University of Science and Technology}\\
Hong Kong Special Administrative Region, China}
 \IEEEauthorblockA{$^{2}$\textit{Beijing Laiye Network Technology Co., Ltd} \\
Beijing, China}
 \IEEEauthorblockA{Email: zwanggy@connect.ust.hk}
}

\maketitle

\begin{abstract}
Legal case retrieval, which aims to retrieve relevant cases given a query case, plays an essential role in the legal system. While recent research efforts improve the performance of traditional ad-hoc retrieval models, legal case retrieval is still challenging since queries are legal cases, which contain hundreds of tokens. Legal cases are much longer and more complicated than keywords queries. Apart from that, the definition of legal relevance is beyond the general definition. In addition to general topical relevance, the relevant cases also involve similar situations and legal elements, which can support the judgment of the current case. In this paper, we propose an interaction-focused network for legal case retrieval with a multi-view contrastive learning objective. The contrastive learning views, including case-view and element-view, aim to overcome the above challenges. The case-view contrastive learning minimizes the hidden space distance between relevant legal case representations produced by a pre-trained language model (PLM) encoder. The element-view builds positive and negative instances by changing legal elements of cases to help the network better compute legal relevance. To achieve this, we employ a legal element knowledge-aware indicator to detect legal elements of cases. We conduct extensive experiments on the benchmark of relevant case retrieval. Evaluation results indicate our proposed method obtains significant improvement over the existing methods.
\end{abstract}


\section{Introduction}
\blfootnote{*The work is partially done during internship at Beijing Laiye Network Technology Co., Ltd}
Legal case retrieval, a specialized information retrieval task, has drawn increasing attention rapidly, as legal artificial intelligence can liberate legal professionals from a maze of paperwork~\cite{zhong2020does}. 
%This task differs from the general information retrieval task. 
In countries with a Common Law system, like the United States, searching previous relevant cases is essential since judicial judgments are made according to past precedential judgments of relevant cases. In countries with a Civil Law system, professionals could also use previous similar cases as references. With the rapid growth of digitized legal documents, it takes great effort to search for relevant cases manually. The development of natural language processing (NLP) and information retrieval (IR) has brought new opportunities to the law system for searching in a deluge of legal materials. Legal case retrieval is the research problem about applying NLP and IR techniques to searching legal domain documents.

Many works~\cite{monroy2013link, raghav2016analyzing,  bhattacharya2020hier} recently tried to build legal retrieval engines using links or citations between case documents and legal statutes. Since the recent success of BERT~\cite{devlin2018bert}, pre-trained language models have drawn great attention in the field of legal case retrieval. The BERT-PLI model~\cite{shao2020bert}, a BERT-based model with paragraph-level interactions, is designed for legal documents. However, it suffers from the lack of long-distance attention and limits the model performance~\cite{ding2020cogltx} due to breaking text into paragraphs. At the same time, many models were proposed for the general IR task, but most of these efforts~\cite{guo2016deep, xiong2017end, mitra2017learning} focused on matching between short and long texts, which will cause complexity issues when being applied between long texts. There is little literature for finding methods to handle long text matching~\cite{jiang2019semantic, yang2020beyond}. Thus, matching between long document pairs, which the legal case retrieval belongs to, was less explored, requiring more research work.

%% In particular, precedents that are priorly decided cases are primary legal materials in the law system. 
\begin{table}[t]
	\centering
	\begin{tabular}{p{8cm}}
		\hline
		\textbf{Case A:} In 2010, PersonX borrowed 10k dollars from PersonY at \textbf{10\%} interest rate. PersonY urged the PersonX for the principal and interest, but the PersonX has not paid back so far.\\
		\hline
		\textbf{Case B:} In 2010, PersonX borrowed 10k dollars from PersonY at \textbf{40\%} interest rate. PersonY urged the PersonX for the principal and interest, but the PersonX has not paid back so far.\\
		\hline
		Different legal element: different interest rates (10\% vs. 40\%)\\
		\hline 
	\end{tabular}
	\caption{An example of legal element. Even though the two cases are similar, they are irrelevant in legal case retrieval due to the legal element: different rates of interest.}
	\label{legal element example}
\end{table}

% \begin{table}[t]
% 	\centering
% 	\begin{tabular}{p{8cm}}
% 		\hline
% 		\textbf{Legal case:} This court affirmed that the debtor-creditor relationship between Company X and Y Village Committee due to the transfer of creditor’s rights was legal and valid.\\
% 		\hline
% 		Legal element: Transfer of creditor’s rights\\
% 		\hline
% 		\textbf{Legal case:} The court believes that the debtor-creditor relationship between X village committee and Y bank is real, legal, and effective.\\
% 		\hline
% 		Legal element: None\\
% 		\hline 
% 	\end{tabular}
% 	\caption{The examples of legal elements in legal cases.}
% 	\label{legal element example}
% \end{table}
Whether designed for the general or legal domain, existing methods still face a few main challenges when applied to legal case retrieval~\cite{shao2020bert}: 

(1) both the query and candidate cases involve long texts. Cases in CAIL2019-SCM, a legal case retrieval dataset, contain around 650 words on average. Long texts cause several issues~\cite{yang2020beyond}, including superficial text understanding and high memory consumption.

(2) the definition of legal domain relevance is beyond the general definition. Relevant cases are not only related to content but also can support judgments of each other~\cite{van2017concept}. General IR models only capture whether two pieces of texts contain similar content but fail to consider their judgments~\cite{shao2020bert}. 

Legal elements are crucial situations in a case (e.g. interest rate, whether debit note exists, whether borrowers refuse to repay the debt, and so on). They can help the model understand legal cases' judgments well~\cite{shao2020bert}. They also play a pivotal role in affecting experts' judgment of legal relevance in manual search. An example of legal element is shown in Table~\ref{legal element example}. In general IR, the two cases are considered as relevant texts due to many shared words and phrases. In legal case retrieval, they are irrelevant and have totally different judgments because Case B has a surprisingly high rate of interest and becomes usury. We can observe that legal elements have a significant impact on judgments and legal relevance. In practice, even though they are essential for the definition of legal relevance, it is difficult to model legal elements in current neural networks~\cite{zhong2020does}.
%%Another relatively trivial issue~\cite{shao2020bert} is that building up a large dataset requires a great deal of legal expertise and is restricted in many law systems, but it is out of the scope of this paper.
%% As the final challenges, building up a large dataset requires a great deal of legal expertise by hiring professionals and significantly increase costs. Beyond that, collecting a large quantity of legal documents is restricted or forbidden in many law systems. 

To tackle the above challenges, we propose an interaction-focused network with a \textbf{M}ulti-\textbf{V}iew \textbf{C}ontrastive \textbf{L}earning objective. The whole model is entitled MVCL. The proposed model builds word-level interactions based on pre-trained language models (PLMs) and then aggregates these interactions to get the matching representation between a query and a candidate case. We also introduce a multi-view contrastive learning objective, including case-view and element-view, to tackle the challenges mentioned earlier. 
%% Specifically, we utilize attention-pooling to pool representations produced by the PLM encoder and pull closer representations of relevant cases using multi-view contrastive learning.  while most previous works sought to improve the limit of modeling capacity
The case-view contrastive learning can help the model understand long texts more thoroughly (Challenge 1). The element-view contrastive learning builds positive and negative instances by changing legal elements of cases. In this way, legal elements are integrated into our neural network. We also introduce a legal element knowledge-aware indicator to detect elements in legal cases. Through legal element-oriented modeling (i.e., element-view contrastive learning), the model can better understand the query and candidate cases, decide whether two cases' judgments can support each other, and compute legal relevance (Challenge 2). Therefore, the MVCL model can learn legal relevance instead of the general definition.

To evaluate our model, we conduct experiments with benchmarks of legal case retrieval in CAIL2019-SCM. We achieved state-of-the-art results in terms of automatic evaluation. In summary, our main contributions include: 
\begin{itemize}
\item We propose a multi-view contrastive learning objective for legal case retrieval, which can effectively use the cases and elements to align representations of legal cases.

\item We first introduce a legal element knowledge-aware indicator as an external knowledge module of legal elements to improve the model's understanding of legal relevance.

\item  We empirically verify the effectiveness of MVCL on the CAIL2019-SCM dataset and surpass 711 teams in the China AI\&Law Challenge\footnote{http://cail.cipsc.org.cn:2019/}.
\end{itemize}

\section{Related Work}
\subsection{Traditional Ad-hoc Retrival}
Ad-hoc retrieval has received a lot of attention in the past few decades. Traditional models are built on the bag-of-words model and measure the relevance between texts by comparing tokens in texts, including BM25~\cite{robertson1994some} and VSM~\cite{salton1988term}. Inevitably, they face the problem of sparse and high-dimensional representation.

The recent development of deep learning has also inspired applications of neural models in IR. Typically, most approaches can be categorized into two types: representation-focused and interaction-focused~\cite{guo2020deep}. To measure the relevance for documents, models that are representation-focused (i.e., Siamese models) learn the representations of a query and a document separately and then compare two representations by computing distance with functions like cosine, dot product, bilinear, or Euclidean distance. The advantage of these approaches is that shared parameters make Siamese architecture smaller and far more efficient. Nevertheless, representing an entire text using a single vector is not sufficient to capture all crucial information. On the other hand,  interaction-focused approaches first compute a query-document word-by-word similarity matrix to build local interactions and then aggregate these word-level matching results to generate the final relevance score. Well-known methods include MatchPyramid~\cite{pang2016text}, Match-SRNN~\cite{wan2016match}, DRMM~\cite{guo2016deep}, and HCRN~\cite{tay2018hermitian}. These models get better performance in traditional ad-hoc retrieval tasks due to the detailed comparison of queries and documents. Additionally, there are neural models that combine representation-focused and interaction-focused approaches~\cite{mitra2017learning, yu2018modelling}.

Regarding the length of query and document, those works mentioned above mainly concentrate on matching between short texts, or between short and long texts. When adopting to matching between long texts, they don't work well and lead to huge memory cost. Matching between long texts was less explored. In this line of work, SMASH RNN~\cite{jiang2019semantic} is proposed to learn representation using a hierarchical architecture of RNN to model multiple abstraction levels of the document structure, including words, sentences, and paragraphs.~\citeauthor{yang2020beyond}~\cite{yang2020beyond} proposed a Transformer-based hierarchical architecture to better model document structure and to increase maximum input text length of general Transformer-based models.

\subsection{Legal Case Retrieval}
Legal case retrieval is still a challenging task, due to its concept of relevance~\cite{van2017concept}, professional terms and expressions, text length~\cite{shao2020bert} and logical structure behind natural languages~\cite{turtle1995text}. Many works tried to build a retrieval engine using links or citations between case documents and legal statutes~\cite{monroy2013link, raghav2016analyzing, bhattacharya2020hier}. Recently,~\citeauthor{shao2021investigating}~\cite{shao2021investigating} and~\citeauthor{liu2021conversational}~\cite{liu2021conversational} studied user behaviors and practical search interactions between users and systems and observed significant differences between legal case retrieval and general web search.

To attain the purpose of computing semantic content, deep learning methods has been applied to Legal Case Retrieval.~\citeauthor{tran2019building}~\cite{tran2019building} builds an CNN-based model, which also utilizes summarization information and lexical features.~\citeauthor{zhong2020does}~\cite{zhong2020does} adopted a few deep learning models to get a better view of the current progress of Legal Case Retrieval, like BiDAF~\cite{seo2016bidirectional}, SMASH RNN~\cite{jiang2019semantic} and so on.~\citeauthor{shao2020bert}~\cite{shao2020bert} design a BERT-based model with paragraph-level interactions called BERT-PLI, which achieved the state-of-the-art results on the COLIEE 2019 dataset~\cite{rabelo2019summary}. However, it breaks text into paragraphs and suffers from the lack of long-distance attention~\cite{ding2020cogltx}. Thus, it is still worth investigating how to design a BERT-based model to compute the relevance between long case texts.

\begin{figure*}[t]
	\centering 
	\includegraphics[height=8cm]{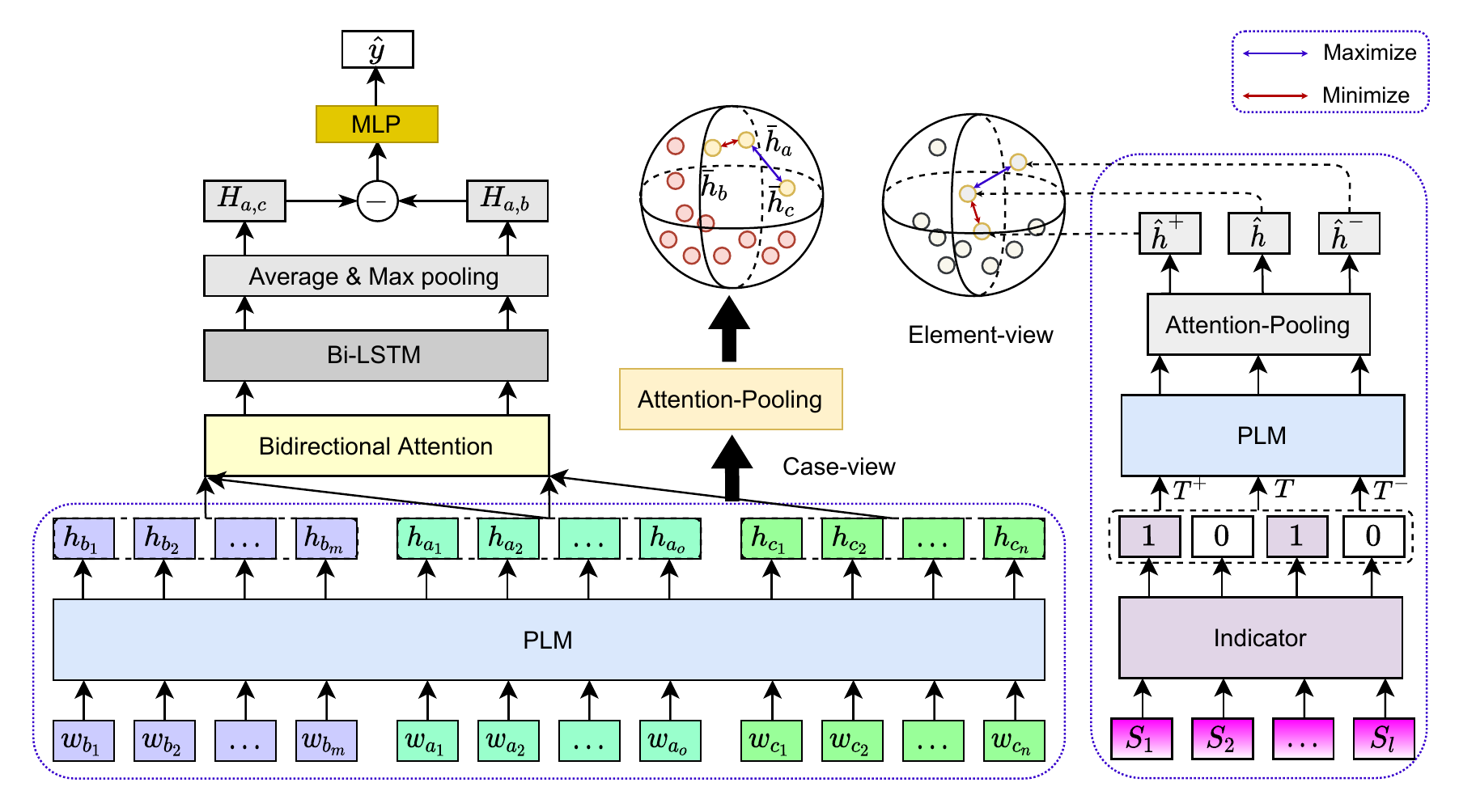}
	\caption{The MVCL model architecture. We divide our model into three parts: (1) The word-level matching module that takes tuples of hidden states $(h_a, h_b)$ and $(h_a, h_c)$ as input, respectively; (2) The case-view contrastive learning with hidden states $\bar{h}_a, \bar{h}_b$ and $\bar{h}_c$; (3) The element-view contrastive learning with selecting sentences ($s_i$ denotes the $i$-th sentence). Note the two PLM blocks mean the same PLM. The details of Indicator is discussed in Subsection~\ref{legal_element_knowledge_aware_indicator}.} 
	\label{legal_model}
\end{figure*}

\subsection{Legal Element Detection}
In addition to common symbols used in information extraction(e.g., people, place, relation), the legal text contains its unique symbols called legal elements. Legal element detection is about detecting crucial situations in a case that affect the decision of the case, like the interest rate in a private lending case.~\citeauthor{CAIL2019-FE}~\cite{CAIL2019-FE} builds a dataset for legal element detection, containing three different kinds of cases (i.e., divorce dispute, labor dispute, and loan dispute). Classical encoding models in NLP like TextCNN~\cite{kim-2014-convolutional} and BERT are implemented on this dataset~\cite{zhong2020does}. The result shows existing models can achieve a promising performance on this dataset, but how to use those elements as domain knowledge to help downstream tasks is still valuable for further exploration. In our approach, we formulate this task as a binary classification task to better help legal case retrieval.

\section{Approach}
\subsection{Problem Formulation}
\label{problem_formulation}
Before presenting our approach for legal case retrieval, we first introduce our problem formulation and notations. The problem is formed as a pairwise ranking task. We take a triple of legal cases $(A,\ B,\ C)$ (e.g., three private lending cases) as input, where $A$ is the query and $(B, C)$ is a pair of candidates. $A = (w_{a_1}, w_{a_2}, ..., w_{a_o}) $, $B = (w_{b_1}, w_{b_2}, ..., w_{b_m}) $ and $C = (w_{c_1}, w_{c_2}, ..., w_{c_n}) $, where $w_{a_i}, w_{b_j}, w_{c_k}$ are tokens and $o, m,$ and $n$ are the number of tokens in $A, B,$ and $C$, respectively. The goal is to determine whether $B$ or $C$ can better support the judicial decision of $A$ (i.e. more relevant at legal level). The label $y$ is a binary variable, where $y=0$ denotes that $B$ is more relevant to $A$ than $C$ and vice versa. After we get pairwise preferences, many methods can convert those preferences into a ranked list of candidates(see examples in Appendix~\ref{converting_methods}). We don't discuss those converting methods in this paper for simplicity.

We also formulate an auxiliary task entitled legal element detection, since we train a model to detect legal elements in sentences (Subsection~\ref{legal_element_knowledge_aware_indicator}) and detection results is utilized to build positive instances in element-view contrastive learning (Subsection~\ref{contrastive_learning}). This task is formed as a binary classification problem. Given a sentence $S$, a legal elements classifier predicts the label $y \in \{0, 1\}$, where $y = 1$ means the sentence contains legal elements.

\subsection{Interaction-focused Network Architecture}
\label{matching_arch}
To begin with, we employ a pre-trained language model as the basic encoder to compute the representations of all words. We will compare different pre-trained language models (PLMs), e.g. Bert, Roberta and XLNet, in the experiment section~\ref{experiments}. This basic layer encodes each case documents independently as the following:

\begin{align}
    \begin{aligned}
    h_a &= {\rm PLM}(w_{a_1}, w_{a_2}, ..., w_{a_o}), \\
    h_b &= {\rm PLM}(w_{b_1}, w_{b_2}, ..., w_{b_m}), \\
    h_c &= {\rm PLM}(w_{c_1}, w_{c_2}, ..., w_{c_n}), 
    \end{aligned}
    \label{PLM}
\end{align}
where the output $h_a, h_b$ and $h_c$ denote the representations for all input words $w_a, w_b$ and $w_c$ in cases $A, B$ and $C$. A word-level matching module is utilized to match representations of a query case and those of a candidate case. Then, an LSTM layer aggregates the matching results to predict a relevant score, following the ESIM framework~\cite{chen2017enhanced}. 

Note that even though the formulation of legal case retrieval indicates a triple $(A,\ B,\ C)$ as input, our approach only needs to compute relevance score between a query and a candidate, like query $A$ and candidate $B$. We can get the relevance between query $A$ and candidate $C$ in the same way. Thus, we only discuss how to compute the relevance score between $A$ and $B$ further on in this paper.   

To better compute legal domain relevance, we first adopt a bidirectional attention layer for query $A$ and candidate $B$, which can easily identify word-level similarities between representations $h_a$ and $h_b$. Specifically, it first computes scaled dot product of $h_{a_i}$ and $h_{b_j}$ as shown in Equation~\ref{dotproduct_att}, where $d$ is the dimension of $h_{a_i}$ and $h_{b_j}$. Then, the scaled dot product is normalized to get attention weights. For a word in query $A$ (i.e. $h_{a_i}$), similar words in the candidate $B$ is identified and aggregated using $\alpha_{{a_i}{b_j}}$, as shown in Equation~\ref{dotproduct_att}, where $\tilde{h}_{{a_i}}$ is the weighted summation of $h_b$. Intuitively, the semantic information relevant to $h_{a_i}$ in $h_b$ will be selected and represented as $\tilde{h}_{{a_i}}$. The same operation is also performed for words in candidate $B$.
\begin{equation}
    \begin{gathered}
    e_{ij} = \frac{h^T_{a_i} h_{b_j}}{\sqrt{d}}, \\
    \alpha_{{a_i}{b_j}} = \frac{\exp(e_{ij})}{\sum_{k=1}^m \exp(e_{ik})},\ \alpha_{{b_j}{a_i}} = \frac{\exp(e_{ij})}{\sum_{k=1}^o \exp(e_{kj})}, \\
\tilde{h}_{{a_i}} = \sum_{j=1}^{m}\alpha_{{a_i}{b_j}} h_{b_j} , \forall i \in \{1,...,o\}, \\
\tilde{h}_{{b_j}} = \sum_{i=1}^{o}\alpha_{{b_j}{a_i}}h_{a_i} , \forall j \in \{1,...,m\},
    \end{gathered}
    \label{dotproduct_att}
\end{equation}
%     \end{gathered}
%     \label{dotproduct_att}
% \end{align}

% \begin{align}
%     \begin{aligned}

Instead of simply concatenating PLM representation (i.e., $h_{a_i}$ or $h_{b_j}$) and weighted summation (i.e., $\tilde{h}_{{a_i}}$ or $\tilde{h}_{{b_j}}$), we further strengthen the local matching information, which is shown helpful in~\cite{chen2017enhanced}. The representations of tuples $\langle h_{a_i}, \tilde{h}_{{a_i}} \rangle$ and $\langle h_{b_j}, \tilde{h}_{{b_j}} \rangle$ are strengthened by computing the subtraction and the element-wise product, as shown in Equation~\ref{strengthen}. $m_{a_i}$ and $m_{b_j}$ are hereinafter called strengthened representation.

\begin{align}
    \begin{aligned}
    m_{a_i} &= [h_{a_i};\tilde{h}_{a_i};h_{a_i}-\tilde{h}_{a_i};h_{a_i} \odot \tilde{h}_{a_i}], \\
    m_{b_j} &= [h_{b_j};\tilde{h}_{b_j};h_{b_j}-\tilde{h}_{b_j};h_{b_j} \odot \tilde{h}_{b_j}]
    \end{aligned}
    \label{strengthen}
\end{align}

Then, a Bi-LSTM layer\footnote{We tested Bi-LSTM, Bi-GRU, and no RNN module and chose Bi-LSTM for better performance. Details will be discussed in the experiment part.} shared for both query and candidate is employed to aggregate the word-level similarity further. We can get the final matching representation through computing both max-pooling and average-pooling~\cite{chen2017enhanced} and concatenating all the vectors:

\begin{align}
    \begin{aligned}
    v_a = {\rm BiLSTM}(m_a),\ v_b = {\rm BiLSTM}(m_b), \\
    H_{a,b} = [v_{a, avg};\ v_{a, max};\ v_{b, avg};\ v_{b, max}],
    \end{aligned}
\end{align}

The above architecture computes and encodes relevance between query $A$ and candidate $B$ into $H_{a, b}$. Following the same way, we can obtain $H_{a, c}$. Finally, the subtraction between $H_{a, b}$ and $H_{a, c}$ is fed into the final multi-layer perceptron (MLP) classifier (shown in Equation~\ref{MLP_and_mainloss}). The MLP classifier contains a hidden layer and a softmax output layer with Tanh activation function between them. We train the model with a cross-entropy loss as follows:

\begin{gather}
    \begin{gathered}
    \hat{y} = {\rm MLP}(H_{a, b} - H_{a, c}) \\
    \mathcal{L} = -(y \log(\hat{y}) + (1 - y) \log(1 - \hat{y}))
    \end{gathered}
    \label{MLP_and_mainloss}
\end{gather}

\subsection{Legal Element Knowledge-aware Indicator}
\label{legal_element_knowledge_aware_indicator}
In this component, we provide the background of the legal element knowledge-aware indicator, which aims to address the auxiliary task mentioned in Subsection~\ref{problem_formulation}. To tackle the Challenge 2 mentioned above, the element-case view contrastive learning (Subsection~\ref{contrastive_learning}) helps our model better focus on legal elements. However, legal element annotations usually are not provided in the current legal case retrieval datasets. This indicator is trained to detect legal elements in those datasets. The annotations are further utilized to build positive instances in element-level contrastive learning. 
%As one of challenges mentioned above, the definition of relevance in the legal scenario is that candidate cases contain legal elements similar to those of query case and consequently support the judicial decision of query case. As shown in Table~\ref{legal element example}, a candidate case with similar words and phrases does not support the judicial decision of the query due to different interest rates (legal elements). 
Our legal element knowledge-aware indicator is basically a binary classifier based on PLM, which is trained to predict whether a sentence contains legal elements.
We input a sentence to PLM\footnote{A few PLMs are tested in the Experiment Section~\ref{experiments}}. Then the indicator outputs a probability distribution of whether the sentence contains legal elements. For autoencoding models (masked language model), like BERT, the hidden state $h_{[CLS]}$ of the special token [CLS] is treated as the sentence representation. For autoregressive models, like XLNet, the representation of the last token is treated as the sentence representation. Then we employ a fully-connected layer with softmax to make the prediction $\hat{y}_s \in \{0, 1\}$, which denotes whether a sentence contains legal elements. The cross-entropy loss is utilized to train the indicator. It is trained independently before training the interaction-focused network mentioned in the Subsection~\ref{matching_arch}.

The indicator can effectively identify whether a sentence contains legal elements. This is important for the matching computation of relevant cases. We use this legal element knowledge-aware indicator as an auxiliary knowledge module to understand cases at the legal elements level.

\subsection{Multi-view Contrastive Learning}
\label{contrastive_learning}

Contrastive learning aims to effectively learn representation by pulling relevant instances together and pushing apart non-relevant ones~\cite{hadsell2006dimensionality}. We are the first to introduce contrastive learning into the legal case retrieval task. We propose multi-view contrastive learning, which contains case view and element view.

% \begin{figure}[t]
% 	\centering 
% 	\includegraphics[height=5cm]{indicator} 
% 	\caption{The legal element knowledge-aware indicator.} 
% 	\label{indicator_pic}
% \end{figure}

\textbf{The case-view contrastive learning} focuses on the input triple $(A, B, C)$ from legal case retrieval datasets. We use the attention-pooling proposed by~\citeauthor{yang2016hierarchical}~\cite{yang2016hierarchical} to pool the $h_a, h_b$ and $h_c$ obtained by Equation~\ref{PLM} as follows:
\begin{align}
    \begin{aligned}
    u_i &= ReLU(Wh_i + b), \\
    \alpha _i &= \frac{\exp(u_i^T u_w)}{\sum_i \exp(u_i^T u_w)}, \\ 
    \bar{h} &= \sum_i{\alpha _i u_i},\ 
    \end{aligned}
\end{align}
where $W$ and $b$ are parameters to transform PLM representations and $u_w$ is a parameter to select words that are important to a case. Moreover, $h_i$ is the PLM representation of the $i$-th word of $A, B$ or $C$ and $\bar{h}$ is the document representation that summarizes all the information in a case. Intuitively, attention-pooling will select important words by assigning them higher attention scores. Prior work~\cite{gunel2020supervised} has demonstrated that supervised datasets are effective for representation learning. In this work, we have the supervised label for query cases. For a triple $(A, B, C)$ with label $B$ from the retrieval dataset, we take case $B$ as the positive instance of $A$ and $C$ as the negative instance, and vice versa for triples with label $C$. Cases from other triples in the same sampled batch are always considered as negative examples. The training objective is defined as follows:

\begin{align}
\begin{gathered}
\mathcal{L}_{case} = -\log\frac{e^{sim(\bar{h},\bar{h}^+)/\mathcal{T}_1}}{e^{sim(\bar{h},\bar{h}^+)/\mathcal{T}_1}+\sum_{\bar{h}^-}e^{sim(\bar{h},\bar{h}^-)/\mathcal{T}_1}}
\end{gathered}
\end{align}
% ,\  sim(\bar{h},\ \bar{h}^-) = \frac{\bar{h}^T\bar{h}^-}{\|\bar{h}\| \|\bar{h}^-\|},
where $\mathcal{T}_1$ is the temperature hyper-parameter and $sim(h_1,h_2)$ is cosine similarity $\frac{h_1^T h_2}{\|h_1\| \|h_2\|}$. Besides, given a triple $(A, B, C)$, $\bar{h}$ is always case $A$. $\bar{h}^+$ and $\bar{h}^-$ are positive and negative instances respectively. The case-view contrastive learning will help the encoder to output similar embedding for similar cases. With case-view contrastive learning, the model focuses more on the important words, like legal elements, and achieves a more thorough understanding of cases.

\textbf{The element-view contrastive learning} directly helps the PLM encoder to focus on legal elements. Consequently, our model can better learn relevance in legal domain. By incorporating this element-view contrastive learning, we pull documents with similar legal elements closer and push documents with different legal elements farther in the representation space. We select a batch of cases from all legal cases that appear in the legal case retrieval dataset uniformly and randomly. We treat a selected case $T$ as a sequence of sentences $S_1, S_2, ..., S_M$. we first use the legal element knowledge-aware indicator to detect whether there are legal elements in every sentence. Thus, we can categorize all sentences into two types: sentences containing legal elements $S_{p_1}, S_{p_2}, ..., S_{p{M_1}}$ and other sentences without legal elements $S_{n_1}, S_{n_2}, ..., S_{n{M_2}}$, where $M_1 + M_2 = M$. 

For constructing positive instances $T^+$, we uniformly and randomly select sentences that don't contain legal elements and replace tokens in those sentences with a new special token [DEL]. Thus, the positive instances share the same legal elements as the original cases. For all deleted sentences $S_{p}^{(1)}, S_{p}^{(2)}, ..., S_{p}^{(q)}$, the total length is $\sum_{i=1}^q l_{p}^i = L_1$, where $l_{p}^i$ is the length of a sentence (The last sentence is potentially truncated) and $L_1$ is a hyper-parameter. Negative instances should contain different legal elements. Randomly sampled cases in the same batch rarely contain the same set of legal elements. Thus, other sampled cases in the same batch are treated as the negative instances. 

After getting positive and negative instances, we first use the same PLM in Equation \ref{PLM} to get hidden states $h$, $h^+$ and $(h_1^-, h_2^-, ..., h_{2N-2}^-)$  for $(T\ , T^+, (T_1^-, T_2^-, ..., T_{2N-2}^-))$, where $N$ is the batch size. Then, we use attention-pooling to pool hidden states $h$, $h^+$ and $(h_1^-, h_2^-, ..., h_{2N-2}^-)$ and get $\hat{h}$, $\hat{h}^+$ and $(\hat{h}_1^-, \hat{h}_2^-, ..., \hat{h}_{2N-2}^-)$, which are document representations for input cases. Given a original case, a positive instance and negative instances $(T\ , T^+, (T_1^-, T_2^-, ..., T_{2N-2}^-))$, we adopt the following loss to compute the element-view training objective:
\begin{equation}
\mathcal{L}_{ele} = -\log\frac{e^{sim(\hat{h},\hat{h}^+)/\mathcal{T}_2}}{e^{sim(\hat{h},\hat{h}^+)/\mathcal{T}_2}+\sum^{2N-2}_{i=1} e^{sim(\hat{h},\hat{h}^-_i)/\mathcal{T}_2}}, 
\end{equation}
where $\mathcal{T}_2$ is temperature hyper-parameter. 
\subsection{Training Objective}
Overall, we first introduced the interaction-focused network architecture, which gives the main objective $\mathcal{L}$. Then, we introduced two contrastive learning objectives: $\mathcal{L}_{case}$ and $\mathcal{L}_{ele}$. The final loss is a weighted sum of these three objectives:
\begin{equation}
    \mathcal{L}_{total} = \mathcal{L} + \lambda_{case}\mathcal{L}_{case} + \lambda_{ele}\mathcal{L}_{ele},
\end{equation}
where $\lambda_{case}$ and $\lambda_{ele}$ are hyper-parameter weights.

\section{Experiments}
\label{experiments}
\subsection{Datasets and Evaluation Metrics}
To verify the effectiveness of MVCL, we employ the benchmark dataset CAIL2019-SCM~\cite{xiao2019cail2019} for legal case retrieval. CAIL2019-SCM contains 8,964 triplets of cases published by the Supreme People’s Court of China, focusing on detecting relevant cases. The participants are required to check which two cases are more relevant in triplets. All cases in CAIL2019-SCM are in the domain of private lending. For the auxiliary task, we employ a legal element extraction dataset CAIL2019-FE~\cite{CAIL2019-FE} to train the legal element knowledge-aware indicator. CAIL2019-FE focuses on extracting elements from three different kinds of cases, including divorce dispute, labor dispute, and loan dispute. We only use the loan dispute data in order to match the domain of CAIL2019-SCM. The statistics are shown in Table~\ref{dataset}. We can see that the average length of cases in CAIL2019-SCM is 676 words, which is longer than texts in most NLP tasks. On the other hand, legal elements are annotated for sentences in CAIL2019-FE. CAIL2019-FE owns regular average length (74 per sentence) because it is for the auxiliary task instead of legal case retrieval.

For dataset selection, there are a few other datasets for the legal case retrieval task, like COLIEE~\cite{kim2018coliee, rabelo2019summary}. However, there are no legal element extraction datasets that focus on the same kind of cases that can be utilized to train an indicator. In this way, other datasets are not suitable to evaluate our approach.

\begin{table}[h]
	\centering
	\begin{tabular}{l|cc}
		\hline
		 &\textbf{CAIL2019-SCM}&\textbf{CAIL2019-FE}\\
		\hline
		\#of training samples    & 5102 & 17567 \\
		\#of validation samples  & 1500 & - \\
        \#of test samples        & 1536 & 5124 \\
        Avg. words       & 676 (per case) & 74 (per sentence)\\

		\hline 
	\end{tabular}
	\caption{The dataset statistics. "CAIL2019-SCM" is the benchmark dataset for legal case retrieval. "CAIL2019-FE" is used for the auxiliary task: learning the Legal Element Knowledge-aware Indicator.}
	\label{dataset}
\end{table}
We adopt accuracy score as an evaluation metric, which is widely applied for legal case retrieval. To better evaluate, we also consider precision, recall, and F1-score to be evaluation metrics. However, compared to common binary classification tasks, our task cares about both positive class (i.e., $y=1$) and negative class (i.e., $y=0$). Thus, we compute macro average precision (MaP), macro average recall (MaR), and macro average F1-score (MaF), instead of directly using precision, recall, and F1-score.

\subsection{Baselines}
To verify the effectiveness of MVCL, we compare our model with four types of baselines: traditional bag-of-words retrieval models, common deep retrieval models, hierarchical models for long text matching, models built for legal case retrieval.

\noindent\textbf{\emph{Traditional bag-of-words retrieval models}} include:
\begin{itemize}
    \item \textbf{TF-IDF:} TF-IDF simply compares terms between query and candidates and also introduces Inverse Document Frequency, which punishes terms that are common in all documents.
    \item \textbf{BM25:} BM25 is another exact matching-based and highly effective method.
\end{itemize}

\textbf{\emph{Common deep retrieval models}} include:
\begin{itemize}
    \item \textbf{MatchPLM:} MatchPLM~\cite{yang2020beyond} uses a PLM to encode both queries and candidates. A dense layer is used to compute similarities between representations of queries and candidates. 
    \item \textbf{MatchPyramid:} MatchPyramid~\cite{pang2016text} is an interaction-focused model, which utilizes a CNN layer to capture matching patterns on a word-level similarity matrix.
    \item \textbf{DRMM:} DRMM~\cite{guo2016deep} also builds a word-level similarity matrix and uses a histogram mapping function. All histograms are transformed by a feed forward network and selected by a gating network to produce the final representation. 
\end{itemize}

\noindent\textbf{\emph{Hierarchical models for long text matching}} include:
\begin{itemize}
    \item \textbf{SMASH RNN:} SMASH RNN~\cite{jiang2019semantic} adopted attentive RNN components in a hierarchical architecture to model multiple abstraction levels, including words, sentences, and paragraphs. 
    \item \textbf{HAN:} HAN~\cite{yang2016hierarchical} builds a hierarchical network based on GRU and uses attention-pooling to aggregate word and sentence representations.
    \item \textbf{SMITH:} SMITH~\cite{yang2020beyond} is another hierarchical model that adopts transformers to compute representations in word and sentence levels.
\end{itemize}
Like MatchBERT, we employ a dense layer as a similarity function for SMASH RNN, HAN, and SMITH.

\noindent\textbf{\emph{Models built for legal case retrieval}} include:
\begin{itemize}
    
    \item \textbf{Roformer:} Roformer~\cite{su2021roformer} is a transformer-based model, which incorporates explicit relative position and is able to model long sequences.
    % \item \textbf{MA PLM:} MA PLM~\cite{MABERT} is one of the top models in the CAIL2019 competition, which builds mutual attention between the query and the candidate based on pretrain language models. 
    \item \textbf{BERT-PLI:} BERT-PLI~\cite{shao2020bert} breaks queries and candidates into paragraphs and computes the relevance between paragraphs. It aggregates paragraph relevance with a RNN layer. BERT-PLI is a strong baseline, which achieves the state-of-the-art on COLIEE 2018’s dataset.
\end{itemize}
\begin{table*}[t]
	\centering
	\begin{tabular}{lcccccccc}
	    \toprule
	    &\multicolumn{4}{c}{\textbf{Validation}} &\multicolumn{4}{c}{\textbf{Test}}\\
		\cmidrule(lr){2-5}\cmidrule(lr){6-9}
		\textbf{Models}&\textbf{Acc.}&\textbf{MaP}&\textbf{MaR}&\textbf{MaF}&\textbf{Acc.}&\textbf{MaP}&\textbf{MaR}&\textbf{MaF}\\
		\hline
		TF-IDF &53.20 & 53.27 & 53.31 & 53.07 &53.45 & 53.45 & 53.46 & 53.43 \\
		BM25 & 54.80& 54.52 & 54.57 & 54.48 &53.77&53.71&53.72&53.71\\
		\hline
		MatchPLM-BERT & 64.93 & 64.81& 65.00 & 64.76 & 65.69&65.69& 65.72& 65.67\\
		MatchPLM-OpenCLaP & 64.73 & 64.58 & 64.77 & 64.55 & 66.60 & 66.57 & 66.61 & 66.57\\
		MatchPLM-Roberta & 65.46 & 65.45 & 65.66 & 65.35 & 66.27 & 66.29 & 66.33 & 66.26 \\
		MatchPLM-XLNet  & 65.93 & 65.73 & 65.93 & 65.73 & 68.16 & 68.12 & 68.15 & 68.12 \\
		MatchPyramid & 63.73 & 63.68 & 63.86 & 63.59 & 68.09 & 68.03 & 68.03 & 68.03\\
		DRMM & 58.20 & 58.23 & 58.34 & 58.07 & 55.59 & 55.59 & 55.60 & 55.57\\
	    \hline
        SMASH RNN & 65.06 & 64.92 & 65.12 & 64.89 &66.27& 66.24& 66.27& 66.24\\
        HAN & 64.87 & 64.77 & 64.97 & 64.71 & 67.31 & 67.29 & 67.33 & 67.29\\
        SMITH-BERT & 62.06 & 62.03 & 62.19 & 61.92 & 63.54 & 63.46 & 63.45 & 63.46 \\
        SMITH-OpenCLaP & 63.80 & 63.81 & 64.00 & 63.68 & 62.82 & 62.86 & 62.88 & 62.81 \\
        SMITH-Roberta & 63.73 & 63.62 & 63.80 & 63.56 & 62.43 & 62.38 & 62.40 & 62.39\\
        SMITH-XLNet & 63.40 & 63.21 & 63.37 & 63.19 & 61.13 & 61.13 & 61.16 & 61.11 \\
		\hline
		Roformer & 66.07 & - & - & - &69.79& -&-&-\\
% 		MA PLM-BERT & 68.06 & 68.02 & 68.26 & 67.94 & 70.57 & 70.54 & 70.57 & 70.54 \\
% 		MA PLM-OpenCLaP & 66.60 & 66.58 & 66.81 & 66.48 & 70.83 & 70.80 & 70.84 & 70.80 \\
% 		MA PLM-Roberta & 68.60 & 68.76 & 69.01 & 68.53 & 70.96 & 70.92 & 70.95 & 70.93 \\
% 		MA PLM-XLNet & 68.33 & 68.44 & 68.69 & 68.25 & 71.28 & 71.25 & 71.28 & 71.25 \\
		
		BERT-PLI-BERT & 67.20 & 67.19 & 67.42 & 67.08 & 68.61 & 68.55 & 68.54 & 68.55\\
		BERT-PLI-OpenCLaP & 66.00 & 65.99 & 66.21 & 65.88 & 70.11 & 70.10 & 70.14 & 70.09 \\
		BERT-PLI-Roberta & 67.20 & 67.32 & 67.55 & 67.12 & 69.20 & 69.16 & 69.19 & 69.16\\
		BERT-PLI-XLNet & 67.80 & 67.76 & 68.01 & 67.68 & 69.79 & 69.77 & 69.81 & 69.77\\
		\hline
		MVCL-BERT & 68.53 & 68.46 & 68.71 & 68.40 & 70.44
		& 70.41 & 70.45 & 70.41 \\
		MVCL-OpenCLaP & \underline{69.93} & \underline{69.75} & \underline{70.00} & \underline{69.76} &\textbf{73.69}
		&\textbf{73.70}&\textbf{73.75}&\textbf{73.68}\\
		MVCL-Roberta & \textbf{70.00} & \textbf{69.96} & \textbf{70.23} & \textbf{69.88} &71.35
		&71.31&71.34&71.32\\
		MVCL-XLNet & 68.33 & 68.38 & 68.63 & 68.24 & \underline{72.78} & \underline{72.78}
		& \underline{72.83} & \underline{72.77} \\
		\bottomrule
	\end{tabular}
	\caption{Performance of four types of baseline and our model ("MVCL-" rows). '-BERT', '-OpenCLap', '-Roberta' and '-XLNet' denote the PLM used in a model, e.g. 'BERT-PLI-Roberta' means using Roberta in BERT-PLI, instead of BERT. The best performance and second-best performance are highlighted using bold and underline fonts respectively}
	\label{overall}
\end{table*}
\subsection{Experimental Settings}
In the legal documents, in addition to the description of the case, some meta-information is also provided, such as names or pseudonyms of plaintiffs, defendants, and lawyers. However, this meta-information varies with documents, contribute nothing to performance, and causes privacy problems. Thus, we first use regular expressions to match that meta-information and omit those in the description. 

For training the legal element knowledge-aware indicator, PLMs take a sentence as its input with 512 as maximal length. Sentences beyond this length will be truncated from the front. We experimented with a few PLMs as the base module, including BERT\footnote{huggingface.co/bert-base-chinese}~\cite{devlin2018bert}, OpenCLaP\footnote{An autoencoding model pre-trained on about 25 million civil law documents}~\cite{zhong2019openclap}, Roberta\footnote{huggingface.co/hfl/chinese-roberta-wwm-ext}~\cite{liu2019roberta}, and XLNet\footnote{huggingface.co/hfl/chinese-xlnet-base}~\cite{yang2019xlnet}. We will discuss performance of different PLMs in Subsection~\ref{discussion_of_indicator}. Roberta is chosen to initialize the legal element knowledge-aware indicator due to its best empirical performance. We use the Adam optimizer and set the learning rate as $2\times 10^{-5}$. The batch size is set to 32. The indicator converges after 3,000 steps (about 5.5 epochs) and F1-score on the test set reaches $88.96$.

For the matching part, we first augment the train data by exchanging candidates. More specifically, for a triple (A, B, C), we exchange candidate B and candidate C and get another triple (A, C, B). Similar to the indicator, the documents longer than 512 tokens are truncated from the front. In addition to BERT, we also test the OpenCLaP, Roberta, and XLNet as the basic encoder. The loss weight $\lambda_{case}$ and $\lambda_{ele}$ are both set to 0.01 so that they are not predominant in the total loss. We utilize Adam optimizer on mini-batches of size 16 with the initial learning rate $2\times10^{-5}$. We train our model for 3500 (about 5.5 epochs) iterations and use a linear learning rate scheduler with the first 350 steps (10\%) for warmup. The best model is selected in the training process according to the accuracy measure on the validation set.

As for the baseline methods, TF-IDF and BM25 are calculated based on the standard scoring functions. PLM-based models are trained about 5 epochs and other deep learning models are trained about 60 epochs. Learning rates are about $2\times10^{-5}$. More details can be found in Appendix~\ref{baseline_detail}.

\subsection{Overall Performance}
Table~\ref{overall} shows the performance on validation set and test set. Our proposed model shows significant improvement in all metrics. The traditional bag-of-words retrieval models perform much worse than the deep learning models because they can't have the semantic understanding ability of documents and simply do exact comparison. Even though common deep retrieval models perform better, their performance is still unsatisfactory. We notice that MatchPyramid achieves relatively high performance among all baselines, while it's only based on CNN. This may be attributed to the word-level similarity matrix, which introduces fine-grained semantic matching. However, another interaction-focused model, DRMM, achieves the worst performance among deep learning models because it is designed for keyword-based queries. So, simply replacing keyword queries with legal cases gives rise to a huge drop in performance. For MatchPLM, we can observe that using XLNet for the PLM encoder achieves competitive performance and that replacing it with OpenCLap, which is pre-trained on civil law documents, can't achieve performance improvement. These observations show that pre-training language models with general domain corpora doesn't introduce too much domain drift when we fine-tune them on legal domain corpora.

For hierarchical models for long text matching, both SMASH RNN and HAN can achieve similar performance to MatchPLM (no matter which PLM is used), only using RNN as their component. This shows that hierarchical architecture is an effective tool to enhance representation-focused models. Another noteworthy observation is that the transformer-based model SMITH doesn't outperform RNN-based models (SMASH RNN and HAN). For hierarchical transformer architecture, the initialization for transformer at sentence-level (and paragraph-level) is a serious problem. SMITH~\cite{yang2020beyond} pre-trained the sentence-level transformer only on English corpus. When we use this model in languages other than English (Chinese in our case), we will face the lack of open-source pre-trained parameters. We can only use token-level PLM (e.g. BERT, OpenCLaP, Roberta and XLNet) to initialize it, which takes words as input, not sentences.

Among models that are built for legal case retrieval, Roformer, which is pre-trained on the dataset same as BERT, performs better than the MatchPLM-BERT. This shows that relative position encoding introduced in Roformer is very helpful in legal case retrieval. Even though BERT-PLI builds interactions between paragraphs and takes the whole case document into consideration without any truncation, it still can't achieve the best performance since breaking documents into different paragraphs omits long distant dependencies across paragraphs~\cite{ding2020cogltx}. 

Table~\ref{overall} shows that our model achieves 70.00\% and 73.69\% in accuracy on validation set and test set respectively, which has already outperformed all the previous models with absolute improvements of 2.20\% and 3.58\%.

% \begin{table}[t]
% \centering
% \begin{tabular}{lcccc}
%     \toprule
% \textbf{Models} & \textbf{Accuracy} & \textbf{MaP} & \textbf{MaR} & \textbf{MaF}\\ 
%     \hline
% max-pooling  &75.13&75.08&75.10&75.09\\
% mean-pooling &74.15&74.10&74.12&74.11\\
%     \bottomrule
% \end{tabular}
% \caption{Performance of different sentence pooling methods.}
% \label{pooling}
% \end{table}

\subsection{Ablation Study}
To understand the effects of multi-view contrastive learning and important modules of interaction-focused network architecture, we conduct an ablation study by removing each component from our framework, and the results are in Table~\ref{ablation}. First, we remove case-view and element-view from the entire model alternately. As illustrated in Table ~\ref{ablation}, there are drops in performance for removing the case-view (- case-view), element-view (- element-view), or both (- multi-view). This shows that our contrastive learning can help the model to understand the long text better and learn legal relevance challenges. 

We also take off the Bidirectional Attention Layer (- BA), Strengthened Representation (- SR) or LSTM (- LSTM) and test the remaining architecture. Taking one of them off will trigger a drastic drop of in accuracy. The results shows that all components are very important in understanding legal cases and computing their relevance. Finally, we replace the LSTM layer with GRU (- LSTM + GRU), which is used to aggregate word-level similarity information. The empirical results show that GRU harms the performance due its simplicity compared to LSTM.

\begin{table}[H]
\centering
\begin{tabular}{lcccc}
    \toprule
\textbf{Models} & \textbf{Accuracy} & \textbf{MaP} & \textbf{MaR} & \textbf{MaF}\\ 
    \hline
MVCL & \textbf{73.69}
		& \textbf{73.70} & \textbf{73.75} & \textbf{73.68} \\
- case-view  & 72.26 & 72.25 & 72.29 & 72.24 \\
- element-view   & 72.20 & 72.20 & 72.25 & 72.18 \\
- multi-view   &71.67 & 71.63 & 71.67 & 71.64\\
- BA & 64.06 & 64.04 & 64.07 & 64.03 \\
- SR & 69.92 & 69.93 & 69.97 & 69.91 \\
- LSTM  & 72.65 & 72.65 & 72.70 & 72.64 \\
- LSTM + GRU & 72.39 & 72.35 & 72.38 & 72.36 \\
    \bottomrule
\end{tabular}
\caption{Ablation study results. In the first three rows, we ablate multi-view contrastive learning objectives. "-BA" and "-SR" ablate "Bidirectional Attention" and "Strengthened Representation," respectively. We also ablate "LSTM" and replace it with "GRU."}
\label{ablation}
\end{table}

% Beyond the above analysis, we test the effect of attention-pooling in our contrastive learning tasks in Table~\ref{pooling}. We can observe that both using max-pooling and mean-pooling hurt the performance, especially the mean-pooling. We believe this drop is attributed to additional restrictions on the PLM encoder's output. For instance, the useful part of the output is forced to be enlarged in max-pooling. Those extra restrictions may contradict the features that MVCL learns.

\subsection{Discussion of the legal element knowledge-aware Indicator}
\label{discussion_of_indicator}
We trained a legal element knowledge-aware indicator in our entire architecture and used it to detect legal elements. The performance of this indicator will greatly affect the performance of the entire model, as the quality of predicted legal element annotations is essential in the element-view contrastive learning objective. Here, we discuss PLMs used in our indicator. Along with BERT, we also test three other PLMs, including OpenCLaP, Roberta and XLNet. We show the performance of BERT, OpenCLaP, Roberta and XLNet in Table~\ref{indicator}, measured by Precision, Recall, and F1-score. We can observe that Roberta performs better than others in F1-score (88.96). Thus, we adopt Roberta as the PLM in our indicator to finish our retrieval experiments.
\begin{table}[h!]
\centering
\begin{tabular}{lccc}
    \toprule
\textbf{Models} & \textbf{Precision} & \textbf{Recall} & \textbf{F1-score}\\ 
    \hline
BERT & 88.30 & 89.43 & 88.86 \\
OpenCLaP & 87.06 & \textbf{89.85} & 88.43 \\
Roberta & \textbf{90.51} & 87.47 & \textbf{88.96}\\
XLNet & 87.13 & 89.69 & 88.39 \\
    \bottomrule
\end{tabular}
\caption{Performances of different indicators. Different pre-trained language models are utilized with a classification head, including BERT, OpenCLaP, Roberta, and XLNet.}
\label{indicator}
\end{table}

\subsection{Case Study}

In this subsection, we select an example to illustrate that our element-view contrastive learning works. This case is taken from a query-candidate pair and anonymized for privacy (P1, P2, and P3 for people). We record the details of bi-directional attention computed in the word-level matching module of MVCL. After that, we highlight the words that get the highest attention weights. We test MVCL, MVCL without element-view contrastive learning, and BERT-PLI. Table~\ref{case_study} shows snippets of this case document that contain highlighted words. For MVCL, we can see words related to legal elements obtain higher attention weights, including ``money amount", ``interest rate", ``loan receipt" and ``whether defendants have been urged and refused to pay back" (The red font). On the contrary, without element-view contrastive learning, the MVCL and BERT-PLI models have focused on ``defendants' name", ``the time of repayment" and ``I have no money now" (The blue font), which are less important. This indicates that element-view contrastive learning helps the model to focus on the words related to legal elements more than other words.

\begin{table}[h]
    \centering
    \begin{tabular}{|p{0.95\linewidth}|}
    \hline
      \textbf{MVCL:} \textbf{...} the defendants (P2) and (P3) jointly borrowed 20,000 yuan from the plaintiff at the agreed {\color{red}monthly interest rate of 3\%}. On that day, the two defendants issued {\color{red}a receipt to the plaintiff}. The receipt states that (P1) borrowed {\color{red}20,000 yuan in cash} at a monthly interest rate of 3 points. Later, the plaintiff {\color{red}repeatedly urged the defendants} for the principal and interest of the loan, but the two defendants {\color{red}have not paid back} so far \textbf{...}
      \\ \hline
      \textbf{MVCL w/o element-view:} \textbf{...} {\color{blue}the defendants (P2) and (P3)} jointly borrowed 20,000 yuan from the plaintiff at the agreed monthly interest rate of 3\%. On that day, the two defendants {\color{red}issued a receipt} to the plaintiff \textbf{...} Later, the plaintiff repeatedly urged the defendant for the principal and interest of the loan, but the two defendants {\color{red}have not paid back so far, so a lawsuit has been filed}. The defendant (P2) argued that she had no objection to the loan \textbf{...} I have no money now and can't give it at the moment. {\color{blue}The defendant (P3) did not reply} \textbf{...} the plaintiff stated to the court that in {\color{blue}February 2016} \textbf{...}
      \\ \hline
      \textbf{BERT-PLI:} \textbf{...} the plaintiff repeatedly {\color{red}urged the defendants for the principal and interest} of the loan, but the two defendants have not paid back so far, so a lawsuit has been filed \textbf{...} I had a mother-daughter relationship with (P3), and {\color{blue}I used the money}. However, I have paid tens of thousands of interest to the plaintiff. The interest is calculated at 3\%, and I acknowledge that the principal has not been returned, but {\color{blue}I have no money now} and can't give it at the moment \textbf{...} {\color{blue}In conclusion, the plaintiff repeatedly} urged the two defendants to repay {\color{red}20,000 yuan of principal and interest} but failed \textbf{...}
      \\ \hline
    \end{tabular}
    \caption{Examples of the attention by different models. Text in {\color{red}red} denotes the significant information, and text in {\color{blue}blue} indicates the unimportant information for case judgment. Note that the original examples are written in Chinese. They are translated into English here.}
    \label{case_study}
\end{table}

\section{Acknowledgement}
This paper was partially supported by the NSFC Fund (U20B2053) from the National Natural Science Foundation of China.

%\section{Conclusions}
%In this paper, we propose an interaction-focused network with a multi-view contrastive learning objective (MVCL) which aims to better learn the case relevance at the legal level by modeling the case-view and element-view contents. To capture the relationship between cases at the legal level, we utilize a legal element knowledge-aware indicator to assist the main model
%which alleviates the problem that it is difficult to define the relevance of legal cases. 
%Experiments on the legal case retrieval benchmark dataset demonstrate that our method out-perform the existing methods. %We also analyze the effectiveness of some modules in the MVCL model. In the future, we can first extend the pre-training model of legal knowledge to understand the legal case in the particular domain; second, we will explore how to extract legal elements without introducing an external dataset.

\newpage
% Generated by IEEEtranN.bst, version: 1.14 (2015/08/26)

\newpage
\appendices

\section{Training Details For Baselines}
\label{baseline_detail}
The performance of Roformer is copied from its paper. For MatchPyramid, there are three CNN layers and kernel sizes are $3\times3,\ 5\times5$ and $7\times7$. Padding sizes are $1,\ 2$ and $3$. All CNN layers has the stride as 2 and the output channel as 64. For DRMM, instead of using term vectors or IDF vectors, we use pre-trained word embeddings in the gating network to consider semantic content. We split the interval of cosine similarity (i.e. [-1, 1]) into 17 bins, whose lengths are equally 0.125. So, all the bins are [-1, -0.875), [-0.875, -0.75), ..., [0.75, 0.875), [0.875, 1) and [1, 1]. Note that we also take the exact match ([1, 1]) as a bin, following the original paper. SMASH RNN takes LSTM as its component. On the other hand, HAN takes GRU, following the original paper. In SMITH, we use BERT, OpenCLaP, Roberta or XLNet as the initialization of both word-level and sentence-level transformers. The block size is 128. For BERT-PLI, we split text into blocks with 256 words, due to there are no explicit paragraph boundaries. 

For the dataset CAIL2019-FE, The organizer of the China AI\&Law Challenge did not publish the validation set and the test set. We use the train data published in the first stage of the challenge as test data and that in the second stage as train data.
\begin{table}[h]
	\centering
	\begin{tabular}{lcccc}
		\toprule
		\textbf{Models}&\textbf{learning rate}&\textbf{batch size}&\textbf{epoch}\\
		\hline
		MatchPLM-BERT & 2e-5 & 16 & 5\\
		MatchPLM-OpenCLaP & 2e-5 & 16 & 10\\
		MatchPLM-Roberta & 2e-5 & 16 & 10 \\
		MatchPLM-XLNet & 2e-5 & 16 & 15 \\
		MatchPyramid & 2e-5 & 12 & 60\\
		DRMM & 2e-5 & 12 & 60\\
	    \hline
        SMASH RNN & 2e-5 & 12 & 60\\
        HAN & 2e-5 & 12 & 60\\
        SMITH-BERT & 2e-5 & 16 & 10\\
        SMITH-OpenCLaP & 2e-5 & 16 & 15\\
        SMITH-Roberta & 2e-5 & 16 & 15\\
        SMITH-XLNet & 2e-5 & 16 & 10\\
		\hline
% 		MA PLM-BERT & 2e-5 & 16 & 5\\
% 		MA PLM-OpenCLaP  & 2e-5 & 16 & 10\\
% 		MA PLM-Roberta & 2e-5 & 16 & 10\\
% 		MA PLM-XLNet  & 2e-5 & 16 & 5\\
		BERT-PLI-BERT & 1e-5 & 8 & 4\\
		BERT-PLI-OpenCLaP & 1e-5 & 8 & 4\\
		BERT-PLI-Roberta & 1e-5 & 8 & 5\\
		BERT-PLI-XLNet & 2e-5 & 8 & 5\\
		\bottomrule
	\end{tabular}
	\caption{Hyper-parameters for deep learning baselines}
	\label{hyper-parameters_baseline}
\end{table}

\section{Methods to convert pair-wise preferences}
\label{converting_methods}
In this section, we list a few methods used to convert pairwise preferences into a ranked list.
\begin{itemize}
    \item We could search all ranked list exhaustively and pick the list that has the largest proportion of correct pairwise preferences.
    \item We could simply rank all candidates according to how many times a candidate is preferred in the pairwise ranking.
    \item We could sum the probabilities of a candidate in all pairs and rank candidates according to their summations of probabilities.
\end{itemize}


\begin{thebibliography}{38}
	\providecommand{\natexlab}[1]{#1}
	\providecommand{\url}[1]{#1}
	\csname url@samestyle\endcsname
	\providecommand{\newblock}{\relax}
	\providecommand{\bibinfo}[2]{#2}
	\providecommand{\BIBentrySTDinterwordspacing}{\spaceskip=0pt\relax}
	\providecommand{\BIBentryALTinterwordstretchfactor}{4}
	\providecommand{\BIBentryALTinterwordspacing}{\spaceskip=\fontdimen2\font plus
	\BIBentryALTinterwordstretchfactor\fontdimen3\font minus
	  \fontdimen4\font\relax}
	\providecommand{\BIBforeignlanguage}[2]{{%
	\expandafter\ifx\csname l@#1\endcsname\relax
	\typeout{** WARNING: IEEEtranN.bst: No hyphenation pattern has been}%
	\typeout{** loaded for the language `#1'. Using the pattern for}%
	\typeout{** the default language instead.}%
	\else
	\language=\csname l@#1\endcsname
	\fi
	#2}}
	\providecommand{\BIBdecl}{\relax}
	\BIBdecl
	
	\bibitem[Zhong et~al.(2020)Zhong, Xiao, Tu, Zhang, Liu, and Sun]{zhong2020does}
	H.~Zhong, C.~Xiao, C.~Tu, T.~Zhang, Z.~Liu, and M.~Sun, ``How does nlp benefit
	  legal system: A summary of legal artificial intelligence,'' in
	  \emph{Proceedings of the 58th Annual Meeting of the Association for
	  Computational Linguistics}, 2020, pp. 5218--5230.
	
	\bibitem[Monroy et~al.(2013)Monroy, Calvo, Gelbukh, and
	  Pacheco]{monroy2013link}
	A.~L. Monroy, H.~Calvo, A.~Gelbukh, and G.~G. Pacheco, ``Link analysis for
	  representing and retrieving legal information,'' in \emph{International
	  Conference on Intelligent Text Processing and Computational
	  Linguistics}.\hskip 1em plus 0.5em minus 0.4em\relax Springer, 2013, pp.
	  380--393.
	
	\bibitem[Raghav et~al.(2016)Raghav, Reddy, and Reddy]{raghav2016analyzing}
	K.~Raghav, P.~K. Reddy, and V.~B. Reddy, ``Analyzing the extraction of relevant
	  legal judgments using paragraph-level and citation information,''
	  \emph{AI4JCArtificial Intelligence for Justice}, p.~30, 2016.
	
	\bibitem[Bhattacharya et~al.(2020)Bhattacharya, Ghosh, Pal, and
	  Ghosh]{bhattacharya2020hier}
	P.~Bhattacharya, K.~Ghosh, A.~Pal, and S.~Ghosh, ``Hier-spcnet: A legal statute
	  hierarchy-based heterogeneous network for computing legal case document
	  similarity,'' in \emph{Proceedings of the 43rd International ACM SIGIR
	  Conference on Research and Development in Information Retrieval}, 2020, pp.
	  1657--1660.
	
	\bibitem[Devlin et~al.(2019)Devlin, Chang, Lee, and Toutanova]{devlin2018bert}
	J.~Devlin, M.-W. Chang, K.~Lee, and K.~Toutanova, ``Bert: Pre-training of deep
	  bidirectional transformers for language understanding,'' in \emph{NAACL-HLT},
	  2019.
	
	\bibitem[Shao et~al.(2020)Shao, Mao, Liu, Ma, Satoh, Zhang, and
	  Ma]{shao2020bert}
	Y.~Shao, J.~Mao, Y.~Liu, W.~Ma, K.~Satoh, M.~Zhang, and S.~Ma, ``Bert-pli:
	  Modeling paragraph-level interactions for legal case retrieval.'' in
	  \emph{IJCAI}, 2020, pp. 3501--3507.
	
	\bibitem[Ding et~al.(2020)Ding, Zhou, Yang, and Tang]{ding2020cogltx}
	M.~Ding, C.~Zhou, H.~Yang, and J.~Tang, ``Cogltx: Applying bert to long
	  texts,'' \emph{NIPS}, vol.~33, pp. 12\,792--12\,804, 2020.
	
	\bibitem[Guo et~al.(2016)Guo, Fan, Ai, and Croft]{guo2016deep}
	J.~Guo, Y.~Fan, Q.~Ai, and W.~B. Croft, ``A deep relevance matching model for
	  ad-hoc retrieval,'' in \emph{Proceedings of the 25th ACM international
	  conference on information and knowledge management}, 2016, pp. 55--64.
	
	\bibitem[Xiong et~al.(2017)Xiong, Dai, Callan, Liu, and Power]{xiong2017end}
	C.~Xiong, Z.~Dai, J.~Callan, Z.~Liu, and R.~Power, ``End-to-end neural ad-hoc
	  ranking with kernel pooling,'' in \emph{Proceedings of the 40th International
	  ACM SIGIR conference on research and development in information retrieval},
	  2017, pp. 55--64.
	
	\bibitem[Mitra et~al.(2017)Mitra, Diaz, and Craswell]{mitra2017learning}
	B.~Mitra, F.~Diaz, and N.~Craswell, ``Learning to match using local and
	  distributed representations of text for web search,'' in \emph{Proceedings of
	  the 26th International Conference on World Wide Web}, 2017, pp. 1291--1299.
	
	\bibitem[Jiang et~al.(2019)Jiang, Zhang, Li, Bendersky, Golbandi, and
	  Najork]{jiang2019semantic}
	J.-Y. Jiang, M.~Zhang, C.~Li, M.~Bendersky, N.~Golbandi, and M.~Najork,
	  ``Semantic text matching for long-form documents,'' in \emph{The World Wide
	  Web Conference}, 2019, pp. 795--806.
	
	\bibitem[Yang et~al.(2020)Yang, Zhang, Li, Bendersky, and
	  Najork]{yang2020beyond}
	L.~Yang, M.~Zhang, C.~Li, M.~Bendersky, and M.~Najork, ``Beyond 512 tokens:
	  Siamese multi-depth transformer-based hierarchical encoder for long-form
	  document matching,'' in \emph{CIKM}, 2020, pp. 1725--1734.
	
	\bibitem[Van~Opijnen and Santos(2017)]{van2017concept}
	M.~Van~Opijnen and C.~Santos, ``On the concept of relevance in legal
	  information retrieval,'' \emph{Artificial Intelligence and Law}, vol.~25, pp.
	  65--87, 2017.
	
	\bibitem[Robertson and Walker(1994)]{robertson1994some}
	S.~E. Robertson and S.~Walker, ``Some simple effective approximations to the
	  2-poisson model for probabilistic weighted retrieval,'' in
	  \emph{SIGIR}.\hskip 1em plus 0.5em minus 0.4em\relax Springer, 1994, pp.
	  232--241.
	
	\bibitem[Salton and Buckley(1988)]{salton1988term}
	G.~Salton and C.~Buckley, ``Term-weighting approaches in automatic text
	  retrieval,'' \emph{Information processing \& management}, vol.~24, no.~5, pp.
	  513--523, 1988.
	
	\bibitem[Guo et~al.(2020)Guo, Fan, Pang, Yang, Ai, Zamani, Wu, Croft, and
	  Cheng]{guo2020deep}
	J.~Guo, Y.~Fan, L.~Pang, L.~Yang, Q.~Ai, H.~Zamani, C.~Wu, W.~B. Croft, and
	  X.~Cheng, ``A deep look into neural ranking models for information
	  retrieval,'' \emph{Information Processing \& Management}, vol.~57, no.~6, p.
	  102067, 2020.
	
	\bibitem[Pang et~al.(2016)Pang, Lan, Guo, Xu, Wan, and Cheng]{pang2016text}
	L.~Pang, Y.~Lan, J.~Guo, J.~Xu, S.~Wan, and X.~Cheng, ``Text matching as image
	  recognition,'' in \emph{AAAI}, vol.~30, no.~1, 2016.
	
	\bibitem[Wan et~al.(2016)Wan, Lan, Xu, Guo, Pang, and Cheng]{wan2016match}
	S.~Wan, Y.~Lan, J.~Xu, J.~Guo, L.~Pang, and X.~Cheng, ``Match-srnn: Modeling
	  the recursive matching structure with spatial rnn,'' in \emph{Proceedings of
	  the Twenty-Fifth International Joint Conference on Artificial Intelligence},
	  2016, p. 2922–2928.
	
	\bibitem[Tay et~al.(2018)Tay, Luu, and Hui]{tay2018hermitian}
	Y.~Tay, A.~T. Luu, and S.~C. Hui, ``Hermitian co-attention networks for text
	  matching in asymmetrical domains.'' in \emph{IJCAI}, 2018, pp. 4425--4431.
	
	\bibitem[Yu et~al.(2018)Yu, Qiu, Jiang, Huang, Song, Chu, and
	  Chen]{yu2018modelling}
	J.~Yu, M.~Qiu, J.~Jiang, J.~Huang, S.~Song, W.~Chu, and H.~Chen, ``Modelling
	  domain relationships for transfer learning on retrieval-based question
	  answering systems in e-commerce,'' in \emph{Proceedings of the Eleventh ACM
	  International Conference on Web Search and Data Mining}, 2018, pp. 682--690.
	
	\bibitem[Turtle(1995)]{turtle1995text}
	H.~Turtle, ``Text retrieval in the legal world,'' \emph{Artificial Intelligence
	  and Law}, vol.~3, no.~1, pp. 5--54, 1995.
	
	\bibitem[Shao et~al.(2021)Shao, Wu, Liu, Mao, Zhang, and
	  Ma]{shao2021investigating}
	Y.~Shao, Y.~Wu, Y.~Liu, J.~Mao, M.~Zhang, and S.~Ma, ``Investigating user
	  behavior in legal case retrieval,'' pp. 962--972, 2021.
	
	\bibitem[Liu et~al.(2021)Liu, Wu, Liu, Zhang, Shao, Li, Zhang, and
	  Ma]{liu2021conversational}
	B.~Liu, Y.~Wu, Y.~Liu, F.~Zhang, Y.~Shao, C.~Li, M.~Zhang, and S.~Ma,
	  ``Conversational vs traditional: Comparing search behavior and outcome in
	  legal case retrieval,'' in \emph{Proceedings of the 44th International ACM
	  SIGIR Conference on Research and Development in Information Retrieval}, 2021,
	  pp. 1622--1626.
	
	\bibitem[Tran et~al.(2019)Tran, Nguyen, and Satoh]{tran2019building}
	V.~Tran, M.~L. Nguyen, and K.~Satoh, ``Building legal case retrieval systems
	  with lexical matching and summarization using a pre-trained phrase scoring
	  model,'' in \emph{Proceedings of the Seventeenth International Conference on
	  Artificial Intelligence and Law}, 2019, pp. 275--282.
	
	\bibitem[Seo et~al.(2016)Seo, Kembhavi, Farhadi, and
	  Hajishirzi]{seo2016bidirectional}
	M.~Seo, A.~Kembhavi, A.~Farhadi, and H.~Hajishirzi, ``Bidirectional attention
	  flow for machine comprehension,'' \emph{arXiv preprint arXiv:1611.01603},
	  2016.
	
	\bibitem[Rabelo et~al.(2019)Rabelo, Kim, Goebel, Yoshioka, Kano, and
	  Satoh]{rabelo2019summary}
	J.~Rabelo, M.-Y. Kim, R.~Goebel, M.~Yoshioka, Y.~Kano, and K.~Satoh, ``A
	  summary of the coliee 2019 competition,'' in \emph{JSAI International
	  Symposium on Artificial Intelligence}.\hskip 1em plus 0.5em minus 0.4em\relax
	  Springer, 2019, pp. 34--49.
	
	\bibitem[Shu et~al.(2019)Shu, Zhao, Zeng, and Ma]{CAIL2019-FE}
	Y.~Shu, Y.~Zhao, X.~Zeng, and Q.~Ma, ``Cail2019-fe,'' Tech. Rep., 2019.
	
	\bibitem[Kim(2014)]{kim-2014-convolutional}
	Y.~Kim, ``Convolutional neural networks for sentence classification,'' in
	  \emph{Proceedings of the 2014 Conference on Empirical Methods in Natural
	  Language Processing}, 2014, pp. 1746--1751.
	
	\bibitem[Chen et~al.(2017)Chen, Zhu, Ling, Wei, Jiang, and
	  Inkpen]{chen2017enhanced}
	Q.~Chen, X.~Zhu, Z.-H. Ling, S.~Wei, H.~Jiang, and D.~Inkpen, ``Enhanced lstm
	  for natural language inference,'' in \emph{ACL-IJCNLP}, 2017, pp. 1657--1668.
	
	\bibitem[Hadsell et~al.(2006)Hadsell, Chopra, and
	  LeCun]{hadsell2006dimensionality}
	R.~Hadsell, S.~Chopra, and Y.~LeCun, ``Dimensionality reduction by learning an
	  invariant mapping,'' in \emph{2006 IEEE Computer Society Conference on
	  Computer Vision and Pattern Recognition (CVPR'06)}, vol.~2.\hskip 1em plus
	  0.5em minus 0.4em\relax IEEE, 2006, pp. 1735--1742.
	
	\bibitem[Yang et~al.(2016)Yang, Yang, Dyer, He, Smola, and
	  Hovy]{yang2016hierarchical}
	Z.~Yang, D.~Yang, C.~Dyer, X.~He, A.~Smola, and E.~Hovy, ``Hierarchical
	  attention networks for document classification,'' in \emph{Proceedings of the
	  2016 conference of the North American chapter of the association for
	  computational linguistics: human language technologies}, 2016, pp.
	  1480--1489.
	
	\bibitem[Gunel et~al.(2020)Gunel, Du, Conneau, and
	  Stoyanov]{gunel2020supervised}
	B.~Gunel, J.~Du, A.~Conneau, and V.~Stoyanov, ``Supervised contrastive learning
	  for pre-trained language model fine-tuning,'' in \emph{International
	  Conference on Learning Representations}, 2020.
	
	\bibitem[Xiao et~al.(2019)Xiao, Zhong, Guo, Tu, Liu, Sun, Zhang, Han, Hu, Wang,
	  et~al.]{xiao2019cail2019}
	C.~Xiao, H.~Zhong, Z.~Guo, C.~Tu, Z.~Liu, M.~Sun, T.~Zhang, X.~Han, Z.~Hu,
	  H.~Wang \emph{et~al.}, ``Cail2019-scm: A dataset of similar case matching in
	  legal domain,'' \emph{arXiv preprint arXiv:1911.08962}, 2019.
	
	\bibitem[Kim et~al.(2018)Kim, Lu, Rabelo, and Goebel]{kim2018coliee}
	M.-Y. Kim, Y.~Lu, J.~Rabelo, and R.~Goebel, ``Coliee-2018: Evaluation of the
	  competition on case law information extraction and entailment,'' in
	  \emph{Proceedings of the Twelfth International Workshop on Juris-informatics
	  (JURISIN 2018)}, 2018, pp. 105--116.
	
	\bibitem[Su et~al.(2021)Su, Lu, Pan, Wen, and Liu]{su2021roformer}
	J.~Su, Y.~Lu, S.~Pan, B.~Wen, and Y.~Liu, ``Roformer: Enhanced transformer with
	  rotary position embedding,'' \emph{arXiv preprint arXiv:2104.09864}, 2021.
	
	\bibitem[Zhong et~al.(2019)Zhong, Zhang, Liu, and Sun]{zhong2019openclap}
	\BIBentryALTinterwordspacing
	H.~Zhong, Z.~Zhang, Z.~Liu, and M.~Sun, ``Open chinese language pre-trained
	  model zoo,'' Tech. Rep., 2019. [Online]. Available:
	  \url{https://github.com/thunlp/openclap}
	\BIBentrySTDinterwordspacing
	
	\bibitem[Liu et~al.(2019)Liu, Ott, Goyal, Du, Joshi, Chen, Levy, Lewis,
	  Zettlemoyer, and Stoyanov]{liu2019roberta}
	Y.~Liu, M.~Ott, N.~Goyal, J.~Du, M.~Joshi, D.~Chen, O.~Levy, M.~Lewis,
	  L.~Zettlemoyer, and V.~Stoyanov, ``Roberta: A robustly optimized bert
	  pretraining approach,'' \emph{arXiv preprint arXiv:1907.11692}, 2019.
	
	\bibitem[Yang et~al.(2019)Yang, Dai, Yang, Carbonell, Salakhutdinov, and
	  Le]{yang2019xlnet}
	Z.~Yang, Z.~Dai, Y.~Yang, J.~Carbonell, R.~R. Salakhutdinov, and Q.~V. Le,
	  ``Xlnet: Generalized autoregressive pretraining for language understanding,''
	  \emph{Advances in neural information processing systems}, vol.~32, 2019.
	
	\end{thebibliography}
\end{document}